\documentclass[conference]{IEEEtran}

\IEEEoverridecommandlockouts

\IEEEoverridecommandlockouts  % only if you need the funding footnote

\usepackage{cite}
\usepackage{amsmath,amssymb,amsfonts}
\usepackage{algorithmic}
\usepackage{graphicx}
\usepackage{textcomp}
\usepackage{xcolor}
\usepackage{threeparttable}
\def\BibTeX{{\rm B\kern-.05em{\sc i\kern-.025em b}\kern-.08em
    T\kern-.1667em\lower.7ex\hbox{E}\kern-.125emX}}

\usepackage{multirow}
\usepackage[normalem]{ulem}
\useunder{\uline}{\ul}{}
\usepackage[utf8]{inputenc}
\usepackage[T1]{fontenc}
\usepackage{lmodern}
\usepackage{siunitx}
\usepackage[hyphens]{url}
\usepackage{hyperref}
\usepackage{svg}
\usepackage{booktabs}

\usepackage[cachedir=minted-cache]{minted}
\definecolor{LightGray}{gray}{0.95}
\usepackage[table]{xcolor}

\begin{document}

\title{Training event-based neural networks with exact gradients via Differentiable ODE Solving in JAX \\

\thanks{Lukas König is funded by the German Federal Ministry for Economic Affairs and Energy (BMWE) project ESCADE (01MN23004D). David Kappel is funded by the Ministry of Culture and Science of the State of North Rhine-Westphalia under project SAIL (grant no. NW21-059A). The authors gratefully acknowledge the Gauss Centre for Supercomputing e.V. (www.gauss-centre.eu) for funding this project by providing computing time on the GCS Supercomputer JUWELS at Jülich Supercomputing Centre (JSC).}
}

\author{
\IEEEauthorblockN{Lukas K\"onig, Manuel Kuhn, David Kappel}
\IEEEauthorblockA{\textit{CITEC} \\
\textit{Bielefeld University} \\
Bielefeld, Germany \\
\texttt{\{lkoenig13,manuel.kuhn,david.kappel\}@uni-bielefeld.de}}
\and
% \IEEEauthorblockN{Manuel Kuhn}
% \IEEEauthorblockA{\textit{CITEC} \\
% \textit{Bielefeld University} \\
% Bielefeld, Germany \\
% \texttt{manuel.kuhn@uni-bielefeld.de}}
% \and
% \IEEEauthorblockN{David Kappel}
% \IEEEauthorblockA{\textit{CITEC} \\
% \textit{Bielefeld University} \\
% Bielefeld, Germany \\
% \texttt{david.kappel@uni-bielefeld.de}}
% \and
\IEEEauthorblockN{Anand Subramoney}
\IEEEauthorblockA{\textit{Department of Computer Science} \\
\textit{Royal Holloway, University of London} \\
Egham, United Kingdom \\
\texttt{anand.subramoney@rhul.ac.uk}}
}

\maketitle

\begin{abstract}
Existing frameworks for gradient-based training of spiking neural networks face a trade-off: discrete-time methods using surrogate gradients support arbitrary neuron models but introduce gradient bias and constrain spike-time resolution, while continuous-time methods that compute exact gradients require analytical expressions for spike times and state evolution, restricting them to simple neuron types such as Leaky Integrate and Fire (LIF). 
We introduce the Eventax framework, which resolves this trade-off by combining differentiable numerical ODE solvers with event-based spike handling. 
Built in JAX, our frame-work uses Diffrax ODE-solvers to compute gradients that are exact with respect to the forward simulation for any neuron model defined by ODEs . It also provides a simple API where users can specify just the neuron dynamics, spike conditions, and reset rules.
Eventax prioritises modelling flexibility, supporting a wide range of neuron models, loss functions, and network architectures, which can be easily extended.
We demonstrate Eventax on multiple benchmarks, including Yin-Yang and MNIST, using diverse neuron models such as Leaky Integrate-and-fire (LIF), Quadratic Integrate-and-fire (QIF), Exponential integrate-and-fire (EIF), Izhikevich and Event-based Gated Recurrent Unit (EGRU) with both time-to-first-spike and state-based loss functions, demonstrating its utility for prototyping and testing event-based architectures trained with exact gradients.
We also demonstrate the application of this framework for more complex neuron types by implementing a multi-compartment neuron that uses a model of dendritic spikes in human layer 2/3 cortical Pyramidal neurons for computation. Code available at \url{https://github.com/efficient-scalable-machine-learning/eventax}.
% On Yin-Yang, QIF neurons achieve 99.28\% test accuracy with integral loss, while on MNIST with LIF neurons we obtain 97.50 ± 0.07\% test accuracy, consistent with prior EventProp results. 
% Eventax prioritizes modelling flexibility, removing the requirement for closed-form solutions for spike timing or state propagation allowing it to support a wide variety of neurons.
\end{abstract}

\section{Introduction}

Spiking neural networks (SNNs) and, more generally, event-based neural networks (EvNNs), are often formulated in continuous time and compute via discrete spike events.
% enabling temporal coding schemes where precise spike timing carries information. 
Training SNNs with gradient-based methods is challenging because spikes are discontinuous events: gradients must be propagated through both continuous dynamics and discrete resets without degrading temporal precision.
A main trade-off in existing SNN training frameworks is between modelling flexibility and modelling accuracy.
Methods that support arbitrary neuron models typically rely on discretization and surrogate gradients~\cite{lohoff2025snnaxspikingneural, eshraghian2021training, bellec_2018_long}, while methods with exact spike timing \cite{Wunderlich_2021, Klos_2025} are currently limited to specific and analytically tractable neuron types such as LIF.
Our framework, Eventax, aims to resolve this by combining differentiable numerical ODE solving with event-based spike handling. 
This approach yields exact gradients with respect to the forward simulation for any neuron model defined by differential equations, eliminating the need for closed-form solutions for the spike time and state evolution. 
Built on the JAX libraries of Equinox and Diffrax \cite{kidger2021on}, and leveraging the differentiable event-handling machinery introduced by Holberg and Salvi \cite{holberg2024exactgradientsstochasticspiking}, Eventax provides a simple API that lets users define new neuron models by specifying only their dynamics as well as jump- and reset rules. 
Although numerical ODE solving is more computationally intensive than analytical solutions, the resulting modelling flexibility can be valuable for research on complex or biologically inspired neuron dynamics that do not have closed form solutions and for prototyping models for neuromorphic or analog hardware.

\section{Related Work}

\begin{table*}[htbp]
\caption{Comparison of discrete-time and event-based gradient-based SNN optimization libraries.}
\centering
\small
\resizebox{\textwidth}{!}{%
\begin{tabular}{@{}l l l l l l l@{}}
\toprule
\textbf{Library} 
& \textbf{Time scale} 
& \textbf{Backend(s)} 
& \textbf{State evolution}
& \textbf{Spike detection}
& \textbf{Event gradients} 
& \textbf{Supported models}
\\
\midrule
\textbf{Eventax (ours)}      
& continuous
& JAX / Equinox / Diffrax 
& numerical ODE solver
& root-finding
& backprop through solver
& model-agnostic (no closed forms required)
\\
\addlinespace
SNNAX \cite{lohoff2025snnaxspikingneural}
& discrete    
& JAX / Equinox         
& discrete (Euler)
& grid threshold
& surrogate         
& model-agnostic
\\
\addlinespace
Norse \cite{norse2021}
& discrete    
& PyTorch             
& discrete (Euler)
& grid threshold
& surrogate / closed form (LIF)  
& model-agnostic; optional LIF closed form
\\
\addlinespace
snnTorch \cite{eshraghian2021training}
& discrete    
& PyTorch             
& discrete (Euler)
& grid threshold
& surrogate           
& model-agnostic
\\
\addlinespace
Holberg \& Salvi \cite{holberg2024exactgradientsstochasticspiking}
& continuous
& JAX / Equinox / Diffrax
& numerical SDE solver
& root-finding
& backprop through solver
& model-agnostic (single model demonstrated)
\\
\addlinespace
EventProp \cite{Wunderlich_2021}
& continuous
& PyTorch / C++        
& analytical (closed-form)
& root-finding
& closed-form    
& LIF only
\\
\addlinespace
JAXSNN \cite{müller2024jaxsnneventdrivengradientestimation}
& continuous
& JAX                 
& analytical (closed-form)
& direct (Lambert-W)
& closed-form    
& LIF only
\\
\addlinespace
Klos \& Memmesheimer \cite{Klos_2025}
& continuous
& JAX
& analytical (closed-form)
& direct (closed-form)
& autodiff through closed form fwd + pseudodyn.
& LIF \& QIF
\\
\addlinespace
mlGeNN EventProp \cite{shoesmith2025eventproptrainingefficientneuromorphic}
& hybrid 
& GeNN (C++/CUDA)       
& discrete (exp.\ Euler)
& grid threshold
& closed-form
& LIF only (closed-form bwd)
\\
\bottomrule
\end{tabular}
}
\label{tab:libraries}
\end{table*}

There are a variety of software frameworks available for training SNNs using backpropagation (see Table \ref{tab:libraries}). 
These frameworks differ substantially in how they handle event times and compute gradients. 
Broadly, they fall into two categories: discrete-time surrogate-gradient methods and continuous-time event-based methods.

Discrete-time SNN libraries \cite{norse2021, eshraghian2021training, lohoff2025snnaxspikingneural} rely on a fixed time grid and propagate gradients using Backpropagation Through Time (BPTT) with surrogate gradients for the discontinuous spike generation function. 
These frameworks benefit from full auto-diff support, allow for arbitrary neuron and synapse models, and run with high throughput on GPU hardware. 
However, they introduce gradient bias due to the surrogate approximation and limit spike-time precision. 
Usually, this can only be mitigated by increasing the discretization rate, which in turn increases computational and memory costs.

Continuous-time SNN libraries treat spiking neurons as dynamical systems and work on a continuous time scale, restricted only by numerical precision.
However, most existing continuous-time approaches do not integrate the state dynamics numerically. 
Instead, they exploit the fact that certain neuron models like the LIF and specific QIF variants have closed-form analytic solutions for state evolution and spike times. 
This enables event-driven simulation without having to use an ODE-solver to simulate the state over time. %solving ODEs step-by-step. 

For instance, Wunderlich \& Pehle~\cite{Wunderlich_2021} compute the next spike time by root-bracketing and then evaluate the analytic LIF solution.
JAXSNN~\cite{müller2024jaxsnneventdrivengradientestimation}, following G\"oltz et al.~\cite{G_ltz_2021}, uses the Lambert–W function to compute spike times, again relying on closed-form LIF dynamics.
Klos \& Memmersheimer~\cite{Klos_2025} extend this idea to the QIF model, where closed-form trajectories exist under certain constraints. Pseudo-dynamics are added for gradient continuity beyond the trial time.
These methods are efficient and yield gradients which are exact with respect to the underlying continuous-time model, but they lack flexibility: they require closed-form solutions for state integration and spike timing and therefore cannot accommodate arbitrary neuron models.
The MlGENN Eventprop compiler~\cite{shoesmith2025eventproptrainingefficientneuromorphic} employs a hybrid approach. 
It performs discrete Euler updates on a fixed time grid during the forward pass, while using adjoint dynamics during backpropagation to reduce memory consumption.
The closest prior work to ours is that of Holberg et al.\cite{holberg2024exactgradientsstochasticspiking}, who developed the mathematical framework for exact gradients through Event SDEs and contributed the autodifferentiable event-handling solver to Diffrax\cite{kidger2021on}, demonstrating it on a stochastic neuron model. We build on the deterministic subset of this foundation, using Diffrax's differentiable ODE solving and event handling but not the stochastic extensions, to provide a general-purpose SNN framework with abstractions for defining custom neuronal and synaptic dynamics.

Our approach occupies a middle ground between standard BPTT and fully event-based continuous-time methods. 
Like BPTT, we differentiate through the steps of the numerical solver, preserving full end-to-end autodiff and supporting arbitrary neuron dynamics without requiring closed-form solutions.
Like event-based methods, we obtain exact spike times via root-finding, and the solver may take adaptive steps rather than following a fixed grid.
This combination yields exact gradients with respect to the forward numerical computation while remaining agnostic to the choice of neuron model.

\section{Implementation}

Our implementation is built on JAX, combining Equinox and Diffrax to create a differentiable continuous-time SNN framework. Users define neuron models as modular components, then simulate networks using forward functions that return spike times or state trajectories.

\subsection{Forward Functions}

The core of our implementation is the \texttt{EvNN.\_\_call\_\_} method, which performs a single integration step between consecutive events. This function queries the next scheduled event in the spike buffer and integrates the neuron dynamics from the current time until either the next scheduled event or a newly generated spike. If a neuron causes an event during integration, the solver halts, the respective neuron's state is reset, and new events corresponding to its outgoing connections are inserted into the buffer.
Building on this core functionality, the framework provides multiple fully differentiable forward functions for different training objectives:
\begin{itemize}
\item \texttt{EvNN.ttfs} implements time-to-first-spike (TTFS), which simulates the network for given input event times until each output neuron emits one spike or a fixed maximum time window is reached. It returns a vector of first spike times for each output neuron.
\item \texttt{EvNN.state\_at\_t} simulates the network for given input event times until the final observation time, returning the output neurons' states at each observation time as a tensor over times, neurons, and channels.
\end{itemize}

\subsection{Neuron Model Interface} 
Each neuron model \texttt{NeuronModel} specifies its initial state, ODE dynamics $\dot{y} = f(t, y)$, spike condition, input-spike update rule, and reset behaviour (see Fig. \ref{fig:neuron_interface}).

\begin{figure}[htbp]
\centering
\begin{minted}[
    fontsize=\scriptsize,
    linenos,
    numbersep=4pt,
    xleftmargin=1.5em,
    bgcolor=LightGray,
]{python}
class NeuronModel(eqx.Module):
    def init_state(self, n: int) -> State: ...
    def dynamics(self, t: float, y: State) -> State: ...
    def spike_condition(self, t: float, y: State) -> Array: ...
    def input_spike(self, y: State, w: Array) -> State: ...
    def reset_spiked(self, y: State, mask: Array) -> State: ...
\end{minted}
\caption{The \texttt{NeuronModel} interface. Users define custom neuron models by implementing: initial state, dynamics, spike condition, input spike handling, and post-spike reset.}
\label{fig:neuron_interface}
\end{figure}

The framework already accommodates a wide range of models including LIF and QIF, biologically plausible models like the Izhikevich neuron, and continuous-time event-based recurrent units that are machine-learning oriented such as the EGRU, while remaining straightforward to extend with custom models. 
The \texttt{MultiNeuronModel} class enables heterogeneous networks out of previously defined \texttt{NeuronModel} in which different neurons use different dynamical models. 
In addition, the framework provides wrapper classes: functions that take a \texttt{NeuronModel} and return an augmented version with additional behaviour. For example, \texttt{Refractory} extends the neuron with refractory periods, while \texttt{AMOS} restricts neurons to emit At Most One Spike per trial.

\subsection{Neuron Models}

Our framework comes with the following neuron models pre-implemented for convenience

\subsubsection{Leaky Integrate and Fire}
The LIF neuron implemented in Eventax is current based and follows the definition by~\cite{Wunderlich_2021}. 
We have an additional bias current term $I_c$ per neuron (not present in \cite{Wunderlich_2021}) which can be learned.
\begin{equation}
    \tau_{\text{syn}} \frac{\partial I}{\partial t} = -I + I_c
\end{equation}

\subsubsection{Quadratic Integrate and Fire}
The QIF model is implemented as described by~\cite{Klos_2025} in the phase-based version, with a learnable bias current added.
% As in the LIF model, we introduce a learnable bias current.

\subsubsection{Exponential Integrate and Fire}
The EIF model extends the standard LIF with an exponential term that captures the rapid voltage increase near threshold, following the formulation by Fourcaud-Trocm\'{e} et al.~\cite{Fourcaud2003}. 
The membrane dynamics include a soft threshold $v_T$ and slope factor $\Delta_T$ that determine the sharpness of spike initiation. 
While in theory the threshold lies at infinity, to make event detection possible we define spike occurrence at a sufficient cut-off voltage $v_{\text{peak}}$ which is a common practice.

\subsubsection{Izhikevich Model}
We implement the two-variable model introduced by Izhikevich~\cite{izhikevich}, with hyper-parameters $a$, $b$, $c$, and $d$ describing the dynamics.

\subsubsection{Event-Based Gated Recurrent Unit}

We implement the continuous variant of the Event-based Gated Recurrent Unit (EGRU) as described in \cite{subramoney2023efficientrecurrentarchitecturesactivity}. The paper formulates a continuous time version of the EGRU with linear state dynamics similar to the LIF. But, unlike the definition in the paper, we only communicate binary spikes between EGRU-cells which are independent of the units cell state.

\subsubsection{Multi-compartment Dendrite Model}
\begin{figure}[htbp]
    \centering
    \includegraphics[width=0.8\linewidth]{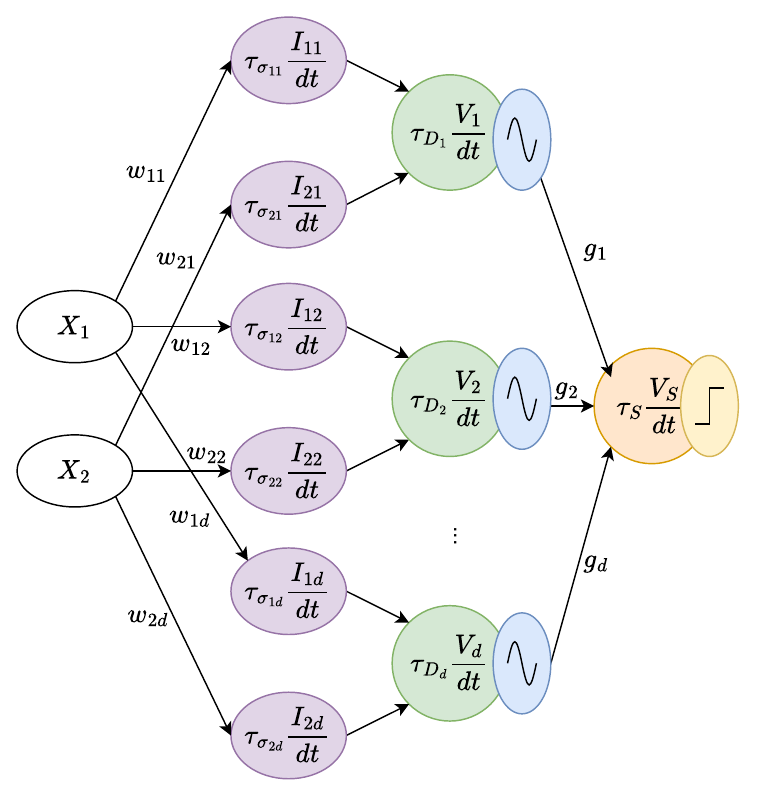}
    \caption{Schematic of the multi-compartment neuron model and corresponding ODEs (Equations~\ref{eq:DLIF:vs_act}-\ref{eq:DLIF:M2}). $X_1$ and $X_2$ are the inputs.}
    \label{fig:dlif}
\end{figure}

Besides neuromorphic machine learning, the framework presented here is also suited for tasks at the intersection of computational neuroscience and machine learning. 
We demonstrate this by implementing and training a multi-compartment neuron model that mimics dendritic spikes observed in human cortical layer 2/3 Pyramidal neurons, to examine the functional properties of dendrites in neuronal information processing.
This model is an extension of the LIF neuron model and adds dendritic compartments to the somatic compartment and voltage dependent dendritic spikes. 
We model a dendritic spike with attenuating amplitude that allows individual dendritic branches to solve non-linear problems in the brain ~\cite{gidonDendriticActionPotentials2020}.
The spike is modelled analogously to~\cite{morrisVoltageOscillationsBarnacle1981} through a voltage dependent activation function $M$. 
The free dynamics of the somatic and dendritic membrane potentials are defined by Equations~\ref{eq:DLIF:vs_act} and~\ref{eq:DLIF:vd}. 
Each synaptic connection $i$ to $d$ has an individual learnable time constant $\tau_{\sigma_{id}}$ as well as every membrane potential. Figure \ref{fig:dlif} shows the structure of the neuron model.
\begin{align}
    \tau_S\dot{v}_S &= - v_S + \sum_{d=1}^D{g_d M(v_{d})(v_S - v_{max})} + I_c \label{eq:DLIF:vs_act}\\
    \tau_{D_d}\dot{v}_{d} &= - v_{d} + \sum_{x_i \in X}{I_{i d}} \label{eq:DLIF:vd} \\
    \tau_{\sigma_{i d}} \dot{I}_{i d} &= -I_{i d} \label{eq:DLIF:i} \\
    M(v) &= \sigma(s_1\,(v-v_{th}))\,\sigma(-s_2\,(v-v_{th}))\label{eq:DLIF:M2}\\\quad &\text{with} \quad \sigma(x) = \frac{1}{1 + e^{-x}} \nonumber
\end{align}

\subsection{Continuous-Time Simulation}

\begin{figure}[htbp]
    \centering
    \includegraphics[width=0.8\linewidth]{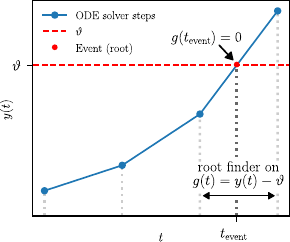}
    \caption{Diffrax event handling. The ODE solver advances the state $y$ until the event condition $g(t, y)$ changes sign. A root-finder then locates the exact event time $t_{\text{event}}$ between the last two solver steps, and the state is integrated from the last step to this exact event time. Gradients are backpropagated directly through the solver steps, while the dependence of $t_{\text{event}}$ on the model parameters is handled via the implicit function theorem.}
    \label{fig:diffrax}
\end{figure}

The simulation evolves neuron states by continuously integrating their ODE dynamics between spike events. Diffrax’s ODE solvers perform this integration while simultaneously monitoring threshold crossings. When a neuron's spike condition changes sign between two solver steps, Diffrax invokes a root-finding procedure to locate the exact spike time and advances the solution to that moment (see Fig \ref{fig:diffrax}).

After an event is detected our framework then applies the model-specific reset rule to the spiking neuron and propagates its effect to all downstream neurons according to the network connectivity.
This approach produces spike times that are exact up to solver accuracy and numerical precision, without requiring closed-form solutions for the dynamics. 
In contrast to discrete-time approximations that detect spikes only at fixed sampling intervals, our continuous-time method captures spikes at exact occurrence times.

\begin{figure}[htbp]
\centering
\begin{minted}[
    fontsize=\scriptsize,
    linenos,
    numbersep=4pt,
    xleftmargin=1.5em,
    bgcolor=LightGray,
]{python}
import jax
import jax.numpy as jnp
from eventpropjax.evnn import FFEvNN
from eventpropjax.neuron_models import QIF
key = jax.random.PRNGKey(0)
max_time = 30.0
net = FFEvNN(
    key=key,
    layers=[20, 3],
    in_size=5,
    neuron_model=AMOS(LIF),
    max_solver_time=max_time,
    tsyn=5.0,
    tmem=20.0,
)
t0, t1, a = 0.5, 6.4, 0.003
def ttfs_loss_single(net, in_spikes, target):
    t = net.ttfs(in_spikes)
    t = jnp.where(jnp.isinf(t), max_time, t)
    logits = -t / t0
    nll = jax.nn.log_softmax(logits)[target]
    reg = jnp.exp(t[target] / t1)
    return -(nll + a * reg)
batched_loss = jax.vmap(ttfs_loss_single, in_axes=(None, 0, 0))
def ttfs_loss(net, in_spikes, targets):
    return batched_loss(net, in_spikes, targets).mean()
loss_and_grad = jax.jit(jax.value_and_grad(ttfs_loss))
for in_spikes, targets in dataloader:
    loss, grads = loss_and_grad(net, in_spikes, targets)
    net = optimizer_step(net, grads)
\end{minted}
\caption{Usage example of EventPropJax: training a LIF network on a TTFS task.  EventPropJax is fully compatible with JAX and therefore works with JAX optimizer libraries, e.g. Optax. We can wrap any neuron with the AMOS wrapper to restrict every neuron to a single spike per trial.}
\end{figure}

\section{Experiments}

\begin{table*}[htbp]
    \centering
    \caption{Neuron model parameters. All models use a fixed ODE solver step size of \SI{0.1}{\ms} (Euler).\\The AMOS wrapper restricts neurons to at most one spike per trial.}
    \label{tab:neuron_params}
    \begin{tabular}{llcccc}
        \toprule
        Parameter & Symbol & LIF & QIF\textsuperscript{a} & EIF\textsuperscript & Izhikevich \\
        \midrule
        Membrane time constant & $\tau_\text{mem}$ & \SI{20.0}{\ms} & \SI{20.00}{\ms} & \SI{20.0}{\ms} & -- \\
        Synaptic time constant & $\tau_\text{syn}$ & \SI{5.00}{\ms} & \SI{5.00}{\ms} & \SI{5.00}{\ms} & \SI{3.0}{\ms} \\
        Spike threshold & $\vartheta$ or $v_\text{peak}$ & 1.00 & $ 1.00 $ & $2.98$ & \SI{30.0}{\mV} \\
        Exponential threshold & $v_T$ & -- & -- & 1.00 & -- \\
        Reset voltage & $v_\text{reset}$ or $c$ & 0.00 & $0.00$ & 0.00 & \SI{-65.00}{\mV} \\
        Leak reversal & $E_L$ & -- & -- & 0.00 & -- \\
        Slope factor & $\Delta_T$ & -- & -- & 0.20 & -- \\
        Recovery time scale & $a$ & -- & -- & -- & 0.020 \\
        Recovery sensitivity & $b$ & -- & -- & -- & 0.20 \\
        Recovery jump & $d$ & -- & -- & -- & 4.00 \\
        \midrule
        $W$ init\textsuperscript{b} & $\mathcal{U}(\mu \pm r)$ & $14 \pm 28$ & $40 \pm 80$ & $20 \pm 40$ & $20 \pm 40$ \\
        $I_c$ init & $\mathcal{U}(\mu \pm r)$ & $0.0025 \pm 0.005$ & $0.0025 \pm 0.005$ & $0.0025 \pm 0.005$ & $3.0 \pm 0.5$ \\
        AMOS (TTFS only) & & \checkmark & -- & -- & \checkmark \\
        \bottomrule
    \end{tabular}
    
    \vspace{0.5em}
    \footnotesize
    \textsuperscript{a} Phase representation $\phi \in [0,1]$; spike at $\phi = 1$, reset to $\phi = 0$~\cite{Wunderlich_2021}. \\
    \textsuperscript{b} Weights scaled by $1/\text{fan-in}$.
\end{table*}

\begin{table}[htbp]
    \centering
    \caption{Training configuration for Yin--Yang and MNIST.}
    \label{tab:training_config}
    \begin{tabular}{lll}
        \toprule
        Parameter & Yin--Yang & MNIST \\
        \midrule
        Network architecture & 5--50--3\textsuperscript{a} & 784--200--10\textsuperscript{a} \\
        Batch size           & 256                        & 1024 \\
        Epochs               & 100                        & 100 \\
        Optimizer            & AdamW\textsuperscript{b} 
                             & AdamW\textsuperscript{b} \\
        Learning rate (TTFS) & 0.005                      & 0.005 \\
        Learning rate (state-based) & 0.005 (LIF) & 0.005 \\
         & 0.03 (others) &  \\
        Gradient clipping (Global norm)    & 1.0            & 1.0 \\
        Seeds per configuration & 5                       & 5 \\
        \bottomrule
    \end{tabular}
    
        \vspace{0.5em}
    \footnotesize
    \textsuperscript{a} fully connected\\
    \textsuperscript{b} with default parameters
\end{table}

In this section we demonstrate Eventax on Yin–Yang~\cite{kriener2022yinyangdataset} and MNIST~\cite{lecun-mnisthandwrittendigit-2010} using both temporal-based and state-based training objectives. 
We train both synaptic weights $W$ as well as bias currents per neuron $I_c$. 
We describe the datasets, network architectures, solvers and training setup, followed by quantitative results.

\subsection{Feed-Forward Network architectures}

For all Yin-Yang and the MNIST experiments we use fully connected feed-forward SNNs implemented via the \texttt{FFEvNN} class. 
Hidden layers consist of $H$ neurons of the chosen neuron model (LIF, QIF, EIF, and Izhikevich).

For temporal-based (TTFS) losses, the output layer uses the same neuron model as the hidden layers, using spike times as the basis of the classification.
For state-based losses, the output layer is composed exclusively of leaky integrator units with an infinite threshold, which integrate but never spike. 
This prevents potential instabilities caused by divergent membrane potentials in models such as QIF and ensures well-behaved logits for the state-based objectives.

\subsection{Input Encodings}
\label{sec:temp_encoding}
We use the Yin–Yang dataset recorded by~\cite{kriener2022yinyangdataset} with 5000 samples for training and 1000 samples each for validation and testing.
Inputs are encoded as five spikes over five input channels based on the $x$ and $y$ coordinate of the Yin-Yang pattern. 
Like~\cite{Wunderlich_2021} we scaled the dataset over $t_{\text{max\_in}} = \SI{30}{\ms}$ and added an additional channel with spike at $t=0$
\begin{equation}
    t_\text{in} =
    \begin{pmatrix}
        0 \\ x \\ y \\ 1-x \\ 1-y
    \end{pmatrix}
    \, \cdot t_{\text{max\_in}},
\end{equation}.

To evaluate scalability to higher-dimensional inputs, we also benchmark on MNIST~\cite{lecun-mnisthandwrittendigit-2010}, a standard handwritten digit classification dataset. We use the standard 60k/10k train/test split, reserving 5k training samples for validation.
Following~\cite{Wunderlich_2021}, each pixel is mapped to an input spike time at an individual channel $k$ with non-zero pixels producing a spike and zero pixels omitted:
\begin{equation}
    t_{\text{in}, k} = 
    \begin{cases} \left(1 - \frac{p_k}{255}\right) \cdot t_\text{max\_in} ~~\text{if} ~~ p_k > 0, \\
    \infty ~~~\text{else}
    \end{cases} 
\end{equation}

\subsection{Temporal-based classification}
\label{sec:ttfs}
For temporal coding, the model predicts the class based on time-to-first-spike (TTFS) among output neurons. 
We adopt the loss from~\cite{Wunderlich_2021}, which applies softmax cross-entropy over negative spike times combined with a regularization term encouraging early target spikes:
\begin{equation}
    \mathcal{L}_{\text{TTFS}} = -\log \frac{e^{-t_y / \tau_0}}{\sum_{c=1}^{C} e^{-t_c / \tau_0}} + \alpha \left( e^{t_y / \tau_1} - 1 \right),
    \label{eq:ttfs_loss}
\end{equation}
where $t_c$ is the first-spike time of output neuron $c$, $y$ is the target class, $\tau_0$ and $\tau_1$ being time constant hyper-parameters, and $\alpha$ balances the two terms.

This objective is fully differentiable in Eventax. 
Spike times are obtained via root finding, with Diffrax applying the implicit function theorem to propagate gradients through event times.

\subsection{State-based classification}

We also evaluate state-based losses following~\cite{Nowotny_2025}. 
These apply softmax cross-entropy to logits derived from output membrane potentials:
\begin{equation}
    \mathcal{L}_{\text{state}} = -\log \frac{e^{z_y}}{\sum_{c=1}^{C} e^{z_c}},
    \label{eq:state_loss}
\end{equation}
where the logits $z_c$ are constructed as:
\begin{align}
    z_c^{\max} &= \max_{t \in [0,T]} V_c(t), \label{eq:logit_max} \\
    z_c^{\text{int}} &= \int_0^T V_c(t)\, dt, \label{eq:logit_int} \\
    z_c^{\exp} &= \int_0^T e^{-\lambda t} V_c(t)\, dt. 
\label{eq:logit_exp}
\end{align}

Here $V_c(t)$ is the membrane potential of output neuron $c$, $T$ is the simulation horizon, and $\lambda = 1/T$ is a temporal decay constant equal to the inverse simulation horizon. 
The max-logit captures peak activation, the integral-logit accumulates total activity, and the exponential-logit emphasizes earlier activity through exponential weighting.

Because models like QIF exhibit diverging potentials, we use Leaky Integrator (LI) neurons (LIF with infinite threshold) in the output layer without bias currents. 
The specified neuron model is used only in the hidden layer.

\subsection{Delayed-memory XOR}
\label{sec:delayed_xor}

To investigate the performance of the continuous EGRU and our models ability to handle recurrent architectures, we implement a delayed-memory XOR task. The network receives nine input channels: the first three encode the binary input value 0, the next three encode the binary input value 1, and the final three encode a cue signal (population \(C\)).

Inputs occur at three time points \(t_0, t_1, t_2\). The first input time is fixed at \(t_0 = 0\). The second input time is drawn uniformly from the early third of the trial,
\begin{equation}
    t_1 \sim \mathcal{U}\!\left(0, \tfrac{1}{3} t_{\max}\right),
\end{equation}
and the cue time is drawn from an interval following the second input,
\begin{equation}
    t_2 \sim \mathcal{U}\!\left(t_1,\, t_1 + \tfrac{1}{3} t_{\max}\right),
\end{equation}
where \(t_{\max}\) denotes the total trial duration. Around each time point
\(t_x \in \{t_0, t_1, t_2\}\), we generate three input spikes by sampling spike times from a uniform distribution
\begin{equation}
    t \sim \mathcal{U}(t_x, t_x + \sigma),
\end{equation}
with \(\sigma\) controlling the temporal jitter.

\begin{figure}[htbp]
    \centering
    \includegraphics[width=0.5\linewidth]{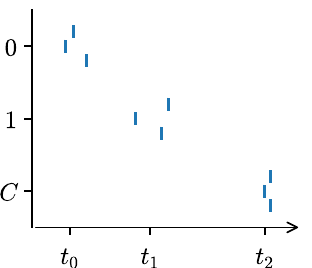}
    \caption{Delayed-XOR task encoding (0, 1)}
    \label{fig:delayed_xor}
\end{figure}

The spikes at \(t_0\) and \(t_1\) are each assigned either to the 0-channels or to the 1-channels, such that the two inputs are either equal (00 or 11) or different (01 or 10). The spikes at \(t_2\) are always assigned to the cue channels \(C\) (see Fig. \ref{fig:delayed_xor}). This construction implements an XOR task over the two temporally separated inputs: trials where the two input populations differ are labelled as class 1, and trials where they are equal as class 0.

For this task, we generate a set of 5000 training samples as well as a test and validation set containing 1000 samples each.

The recurrent layer consists of 32 fully connected neurons without self-connections. Each neuron receives input from all nine input channels. Two leaky integrator output neurons represent the two classes. After the first cue spike, we classify the trial using a softmax over the exponential-integral logit \(z_c^{\exp}\) (Eq.~\eqref{eq:logit_exp}) computed from the output membrane potentials over the remaining simulation interval.

\subsection{Performance Analysis}

We additionally analyse the computational performance of our library. 
For this, we train the QIF model on the Yin–Yang task under a TTFS objective while varying the solver step size. 
For each step size, we retain the model achieving the highest validation accuracy across all epochs. 
Using these selected models, we measure throughput (samples per second) for both forward-only inference and combined forward + backward training passes.

\subsection{Multi-compartment Neuron Model}
We analyse the multi-compartment neuron models by training them for the Yin-Yang task.
The network only consists of one neuron per class which are not interconnected.
Besides the two activations we test the Yin-Yang task for 7 and 14 dendrites. We calculate TTFS directly on the multi-compartment neurons as output neurons, while for the state based losses we append a fully connected output layer of LI-neurons to integrate the state.

\section{Results}

\begin{table*}[htbp]
    \centering
    \begin{threeparttable}
    \caption{Yin-Yang classification with 50 hidden neurons (Except for multi-compartment case, where only 2 neurons were used).\\ Test set performance at best validation epoch (mean $\pm$ std, with median in parentheses) across five seeds trained for 100 epochs.}
    \setlength{\tabcolsep}{3pt}
    \begin{tabular}{llccccc}
        \toprule
        Neuron & Loss & Accuracy [\%] & Max Firing\tnote{a} & Dead Neurons\tnote{b} & Fire Count\tnote{c} & TTFS [\SI{}{\ms}]\tnote{d} \\
        \midrule
        \multirow{4}{*}{LIF}
         & TTFS (AMOS)\tnote{e} & $91.30 \pm 0.94$ (90.90) & $1.00 \pm 0.00$ (1.00) & $2.75 \pm 0.96$ (2.50) & $33.42 \pm 2.00$ (33.06) & $15.82 \pm 0.96$ (15.71) \\
         & Max         & $94.50 \pm 1.21$ (95.20) & $3.94 \pm 0.88$ (3.88) & $9.00 \pm 5.24$ (8.00) & $67.34 \pm 13.65$ (67.93) & -- \\
         & Integral    & $96.68 \pm 0.41$ (96.70) & $3.45 \pm 0.37$ (3.49) & $14.40 \pm 5.81$ (15.00) & $56.87 \pm 11.10$ (58.57) & -- \\
         & Exp-Integral& $97.08 \pm 0.36$ (97.10) & $3.88 \pm 0.09$ (3.88) & $9.00 \pm 3.67$ (7.00) & $63.36 \pm 12.30$ (60.28) & -- \\
        \midrule
        \multirow{4}{*}{QIF}
         & TTFS        & $98.62 \pm 0.31$ (98.70) & $1.00 \pm 0.00$ (1.00) & $1.20 \pm 0.84$ (1.00) & $45.36 \pm 1.35$ (45.43) & $27.27 \pm 0.20$ (27.23) \\
         & Max         & $85.82 \pm 18.10$ (98.60) & $1.38 \pm 0.43$ (1.23) & $0.00 \pm 0.00$ (0.00) & $43.26 \pm 4.04$ (43.06) & -- \\
         & Integral    & $99.28 \pm 0.13$ (99.30) & $1.72 \pm 0.08$ (1.72) & $0.00 \pm 0.00$ (0.00) & $48.83 \pm 0.76$ (48.61) & -- \\
         & Exp-Integral& $99.20 \pm 0.28$ (99.10) & $1.53 \pm 0.31$ (1.63) & $0.00 \pm 0.00$ (0.00) & $47.24 \pm 0.34$ (47.22) & -- \\
        \midrule
        \multirow{4}{*}{EIF}
         & TTFS        & $97.20 \pm 0.54$ (97.20) & $2.53 \pm 0.43$ (2.35) & $1.40 \pm 1.14$ (1.00) & $59.05 \pm 2.95$ (58.37) & $22.00 \pm 1.18$ (21.43) \\
         & Max         & $98.24 \pm 0.38$ (98.10) & $4.16 \pm 0.62$ (3.82) & $1.00 \pm 1.22$ (1.00) & $87.40 \pm 6.89$ (85.38) & -- \\
         & Integral    & $97.90 \pm 0.35$ (97.80) & $3.77 \pm 0.18$ (3.79) & $2.50 \pm 1.29$ (2.50) & $84.37 \pm 5.19$ (84.46) & -- \\
         & Exp-Integral& $97.54 \pm 0.68$ (97.70) & $4.01 \pm 1.02$ (3.68) & $1.00 \pm 1.22$ (1.00) & $91.03 \pm 0.72$ (90.76) & -- \\
        \midrule
        \multirow{4}{*}{Izhikevich}
         & TTFS (AMOS)\tnote{e} & $96.98 \pm 1.53$ (97.10) & $1.00 \pm 0.00$ (1.00) & $0.00 \pm 0.00$ (0.00) & $47.18 \pm 1.34$ (47.31) & $16.77 \pm 0.82$ (16.86) \\
         & Max         & $97.08 \pm 0.99$ (97.70) & $1.75 \pm 0.26$ (1.83) & $4.00 \pm 0.71$ (4.00) & $36.34 \pm 2.14$ (36.94) & -- \\
         & Integral    & $98.20 \pm 0.38$ (98.10) & $1.53 \pm 0.16$ (1.53) & $3.60 \pm 1.52$ (3.00) & $34.36 \pm 3.09$ (32.98) & -- \\
         & Exp-Integral& $98.22 \pm 0.30$ (98.10) & $1.68 \pm 0.26$ (1.75) & $3.40 \pm 1.14$ (3.00) & $34.57 \pm 3.75$ (33.39) & -- \\
        \midrule
        \multirow{4}{*}{Multi-Comp.}
         & TTFS & $88.00 \pm 13.40$ (96.85) & $1.00 \pm 0.00$ (1.00) & $0.00 \pm 0.00$ (0.00) & $1.00 \pm 0.00$ (1.00) & $12.1 \pm 3.63$ (11.84) \\
         & Max         & $93.65 \pm 1.62$ (93.95) & $223,41 \pm 25.79$ (220,46) & $0.00 \pm 0.00$ (0.00) & $518.1 \pm 53.42$ (533.58) & -- \\
         & Integral    & $96.16 \pm 0.94$ (96.45) & $54.57 \pm 20.16$ (51.24) & $0.00 \pm 0.00$ (0.00) & $91.05 \pm 35.78$ (98.06) & -- \\
         & Exp-Integral& $96.50 \pm 0.98$ (96.65) & $57.07 \pm 19.46$ (48.74) & $0.00 \pm 0.00$ (0.00) & $96.19 \pm 34.15$ (98.36) & -- \\
        \bottomrule
    \end{tabular}
    \begin{tablenotes}
        \footnotesize
        \item Spike metrics (a--c) measured until first output spike for TTFS loss, until trial end for state-based losses.
        \item[a] Avg.\ max spike count of any single neuron per test sample.
        \item[b] Neurons with zero spikes across all test samples.
        \item[c] Avg.\ total spikes per test sample.
        \item[d] Avg.\ time to first output spike; ``--'' = not applicable.
        \item[e] Restricted to At Most One Spike (AMOS) per neuron.
    \end{tablenotes}
    \label{tab:yin_yang_results}
    \end{threeparttable}
\end{table*}

\subsection{Yin-Yang Task}
Looking at the results (Table \ref{tab:yin_yang_results}), we can see that our library is able to solve the Yin-Yang task using several different types of neuron models using both state- as well as temporal based classification. 
We can see that models with non-linear dynamics clearly outperform the LIF neuron independent of the loss function used. 
At the same time, the LIF shows the highest amount of dead neurons across all trials. 
This may relate to differences in spike generation dynamics between these neuron models. 
As analysed by Klos et al. \cite{Klos_2025} in the LIF neuron, the voltage derivative $\dot{V}$ at threshold can tend to zero, which means spikes can abruptly appear or disappear mid-trial due to small parameter changes. 
In contrast, the QIF, EIF, and Izhikevich models all exhibit large voltage derivatives near threshold. 
This property suggests that once a neuron approaches threshold, small input changes primarily affect spike timing rather than causing spikes to suddenly appear or disappear mid-trial and instead makes spikes only appear at trial end. 
Such non-disruptive spike behaviour could potentially make it easier for the learning process.

\subsection{MNIST}
We also trained LIF neurons with exponential-integral loss on 
latency-encoded MNIST (784--200--10 architecture), achieving 97.50 $\pm$ 0.07\%  test accuracy across five seeds, consistent with prior Eventprop results \cite{Wunderlich_2021}. 

\subsection{Multicompartment-model}

Looking at Table \ref{tab:yin_yang_results} we can see that the multi-compartment model was able to learn the Yin-Yang task. 
Two of the ten runs had significantly lower accuracies, highlighting the susceptibility of the model to the weight initialization. 
Other than that, it learns the task up to a high accuracy. 
This shows that our implementation is able to work with neuron models having more sophisticated dynamics, allowing us to investigate the ability of active dendrites to solve a non-linear problem.
This usually requires a multi-layer network for accuracies higher than 63.8\% ~\cite{kriener2022yinyangdataset}.

\subsection{Delayed XOR}

We further evaluated Eventax on the delayed-memory XOR task using a continuous-time Event-based GRU (EGRU). A recurrent layer of 32 EGRU units received the event-based inputs and projected to two leaky-integrator output neurons for classification. After the cue onset, we decoded the class using an exponential-integral state-based loss over the remaining simulation interval. The EGRU network solved the task, reaching 100\% test accuracy, indicating that it can reliably store and combine temporally separated inputs over variable delays using only event-based communication. This demonstrates that our implementation not only trains feedforward SNNs, but also successfully handles explicitly recurrent architectures.
 
\subsection{Performance}

\begin{figure*}[htbp]
    \centering
    \includegraphics[width=\textwidth]{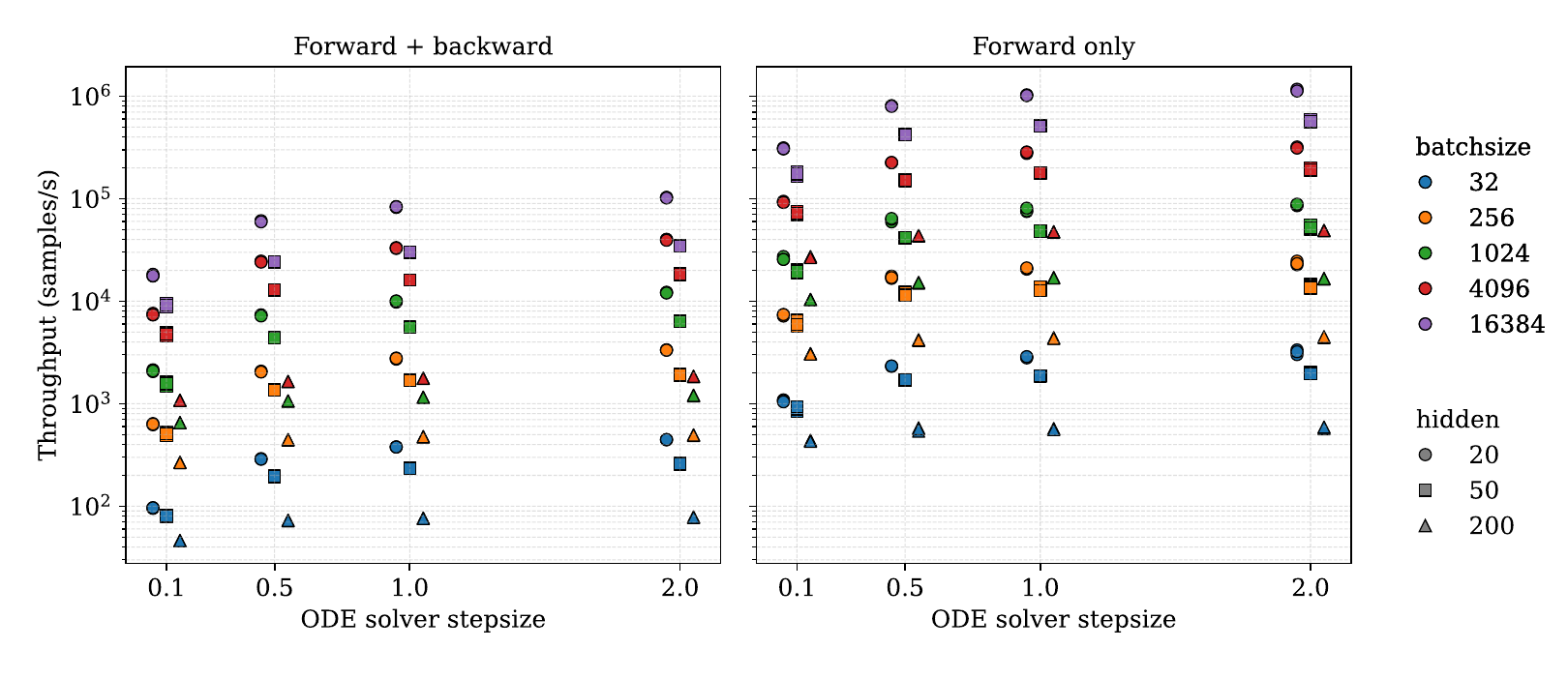}
    \caption{Throughput of QIF models on the TTFS Yin–Yang task (NVIDIA H100). Three models were trained for various Euler step-size and hidden-size combinations, with throughput measured over 50 random batches at each batch size. Results show both forward-only and forward + backward throughput in samples per second.}
    \label{fig:throughput}
\end{figure*}

Our approach has multiple different parameters affecting its speed. 
In Figure \ref{fig:throughput} we can see that because of the usage of Diffrax JAX-capable ODE-solvers, we scale linearly with the batchsize using \texttt{vmap}. 
It is evident that the choice of the ODE solvers stepsize affects the runtime in a similar manner.

We also find that the performance scales with number of events as expected, rather than model size for large step sizes.
For smaller step sizes, the overhead of the ODE solver dominates, with the performance scaling with size of network.

\section{Conclusion and Outlook}

In this paper we present Eventax, a framework that uses Diffrax's differentiable ODE solvers to enable gradient-based training of spiking neural networks with arbitrary neuron models. 
By combining numerical integration with event-based spike handling, Eventax computes gradients that are exact with respect to the forward simulation without requiring closed-form solutions for spike times or state dynamics. 
We demonstrate the framework on the Yin-Yang and MNIST benchmarks using LIF, QIF, EIF, and Izhikevich neurons, achieving competitive accuracies across both temporal and state-based training objectives. 
We also demonstrate the framework on delayed-XOR task using EGRU, demonstrating its applicability to recurrent architectures.
Eventax provides modelling flexibility that may prove valuable for research on biologically inspired neuron dynamics and prototyping models for neuromorphic hardware.

Several promising directions remain for future work. 
First, the event-based formulation is naturally suited to learning synaptic and axonal delays, which may enable networks to exploit richer temporal coding schemes and better align with both biological neural circuits and neuromorphic hardware constraints. 
Second, although Diffrax provides a Backsolve-style adjoint for memory-efficient training, this capability is not yet integrated with event handling; extending Eventax to combine Backsolve with event-based dynamics would yield a complementary low-memory training mode more closely resembling classical event-based backpropagation, albeit at the cost of gradients that are no longer exact with respect to the discretized forward solver. 
Additionally, the dead neuron problem, where neurons cease to spike and thus receive no gradient signal, could be addressed through careful parameter initialization strategies or pseudo-dynamics approaches such as those proposed by Klos et al. \cite{Klos_2025}.

\bibliographystyle{IEEEtran}

% \clearpage

\bibliography{references}

\end{document}